\documentclass[letterpaper]{article}
\usepackage{proceed2e}
\usepackage[margin=1in]{geometry}
\usepackage{graphicx} 
\usepackage{subfigure}
\usepackage[round]{natbib}
\usepackage{algorithm}
\usepackage{algorithmic}
\usepackage{comment}
\usepackage{amsmath,amssymb,amsthm}
\usepackage{hyperref}
\usepackage{multirow,multicol}
\usepackage{algorithm}
\usepackage{algorithmic}

\usepackage{cleveref}
\crefname{equation}{equation}{equations}
\Crefname{equation}{Equation}{Equations}
\crefrangelabelformat{equation}{(#3#1#4--#5#2#6)}
\crefmultiformat{equation}{equations (#2#1#3}{, #2#1#3)}{#2#1#3}{#2#1#3}
\Crefmultiformat{equation}{Equations (#2#1#3}{, #2#1#3)}{#2#1#3}{#2#1#3}

\def\f{\mathbf{f}}
\def\x{\mathbf{x}}
\def\z{\mathbf{z}}
\def\w{\mathbf{w}}
\def\m{\mathbf{m}}
\def\Gamma{\mathrm{Gamma}}

\def\1{\mathbf{1}}
\def\R{\mathbb{R}}
\def\E{\mathbb{E}}

\def\K{\mathbf{K}}

\def\y{\mathbf{y}}

\def\0{\mathbf{0}}
\def\N{\mathcal{N}}
\def\GP{\mathcal{GP}}

\def\bmu{\boldsymbol\mu}
\newcommand{\Var}{\mathrm{Var}}

\renewcommand{\u}{\mathbf{u}}

\newcommand{\g}{{g}}
\newcommand{\bg}{{\mathbf{g}}}

\renewcommand{\L}{\mathcal{L}}

\renewcommand{\S}{{\Sigma}}

\newcommand{\ug}{{\mathbf{u}_g}}

\newcommand{\zf}{{\mathbf{z}_f}}
\newcommand{\zw}{{\mathbf{z}_w}}
\newcommand{\uf}{{\mathbf{u}_f}}
\newcommand{\uw}{{\mathbf{u}_w}}
\newcommand{\mf}{{\mathbf{m}_f}}
\newcommand{\Sf}{{\mathbf{S}_f}}
\newcommand{\mg}{{\mathbf{m}_g}}
\newcommand{\Sg}{{\mathbf{S}_g}}

\newcommand{\pbg}{\Phi(\bg)}

\newcommand{\Sn}{\sigma_y^2}
\newcommand{\fl}{\mathcal{L}}
\newcommand\numberthis{\addtocounter{equation}{1}\tag{\theequation}}
\newcommand{\bk}{\mathbf{k}}

\DeclareMathOperator{\diag}{diag}
\DeclareMathOperator{\KL}{KL}

\DeclareMathOperator{\cov}{\textbf{cov}}

\title{Variational zero-inflated Gaussian processes with sparse kernels}

\author{ {\bf Pashupati Hegde} \And {\bf Markus Heinonen}  \\ \\ Helsinki Institute for Information Technology HIIT \\Department of Computer Science, Aalto University \And {\bf Samuel Kaski}}

\begin{document}

\maketitle



\begin{abstract}
Zero-inflated datasets, which have an excess of zero outputs, are commonly encountered in problems such as climate or rare event modelling. Conventional machine learning approaches tend to overestimate the non-zeros leading to poor performance. We propose a novel model family of \emph{zero-inflated Gaussian processes} (ZiGP) for such zero-inflated datasets, produced by \emph{sparse kernels} through learning a latent probit Gaussian process that can zero out kernel rows and columns whenever the signal is absent. The ZiGPs are particularly useful for making the powerful Gaussian process networks more interpretable. We introduce \emph{sparse GP networks} where variable-order latent modelling is achieved through sparse mixing signals. We derive the non-trivial stochastic variational inference tractably for scalable learning of the sparse kernels in both models. The novel output-sparse approach improves both prediction of zero-inflated data and interpretability of latent mixing models.
\end{abstract}

\section{INTRODUCTION}

Zero-inflated quantitative datasets with overabundance of zero output observations are common in many domains, such as climate and earth sciences \citep{enke1997,wilby1998,charles2004}, ecology \citep{del2008,ancelet2009}, social sciences \citep{bohning1997}, and in count processes \citep{barry2002}. Traditional regression modelling of such data tends to underestimate zeros and overestimate nonzeros \citep{andersen2014}.

A conventional way of forming zero-inflated models is to estimate a mixture of a Bernoulli ``on-off'' process and a Poisson count distribution \citep{johnson1969,lambert1992}. In hurdle models a binary ``on-off'' process determines whether a hurdle is crossed, and the positive responses are governed by a subsequent process \citep{cragg1971,mullahy1986}. The hurdle model is analogous to first performing classification and training a continuous predictor on the positive values only, while the zero-inflated model would regress with all observations. Both stages can be combined for simultaneous classification and regression \citet{abraham2010}.

Gaussian process models have not been proposed for zero-inflated datasets since their posteriors are Gaussian, which are ill-fitted for zero predictions. A suite of Gaussian process models have been proposed for partially related problems, such as mixture models \citep{tresp2001,rasmussen2002,lazaro2012} and change point detection \citep{herlands2016}. Structured spike-and-slab models place smoothly sparse priors over the structured inputs \citep{andersen2014}.

In contrast to other approaches, we propose a Bayesian model that learns the underlying latent prediction function, whose covariance is sparsified through another Gaussian process switching between the `on' and `off' states, resulting in an zero-inflated Gaussian process model. This approach introduces a tendency of predicting exact zeros to Gaussian processes, which is directly useful in datasets with excess zeros.

A Gaussian process network (GPRN) is a latent signal framework where multi-output data are explained through a set of latent signals and mixing weight Gaussian processes \citep{wilson2012}. The standard GPRN tends to have dense mixing that combines all latent signals for all latent outputs. By applying the zero-predicting Gaussian processes to latent mixture models, we introduce sparse GPRNs where latent signals are mixed with sparse instead of dense mixing weight functions. The sparse model induces variable-order mixtures of latent signals resulting in simpler and more interpretable models. We demonstrate both of these properties in our experiments with spatio-temporal and multi-output datasets.

\paragraph{Main contributions.}
Our contributions include
\begin{enumerate}
    \item A novel zero-inflated Gaussian process formalism consisting of a latent Gaussian process and a separate `on-off' probit-linked Gaussian process that can zero out rows and columns of the model covariance. The novel sparse kernel adds to GPs the ability to predict zeros.
    \item Novel stochastic variational inference (SVI) for such sparse probit covariances, which in general are intractable due to having to compute expectations of GP covariances with respect to probit-linked processes. We derive the SVI for learning both of the underlying processes.
    \item A solution to the stochastic variational inference for conventional Gaussian process networks (GPRN) improving the earlier diagonalized mean-field approximation \citep{nguyen13} by taking the covariances fully into account.
    \item A novel sparse GPRN with an on-off process in the mixing matrices leading to sparse and variable-order mixtures of latent signals.
    \item A solution to the stochastic variational inference of sparse GPRN where the SVI is derived for the network of full probit-linked covariances.
\end{enumerate}
The \texttt{TensorFlow} Python implementation of these methods is publicly available at \url{github.com/hegdepashupati/zero-inflated-gp} and at \url{github.com/hegdepashupati/gprn-svi}.

\section{GAUSSIAN PROCESSES}

We begin by introducing the basics of conventional Gaussian processes. Gaussian processes (GP) are a family of non-parametric, non-linear Bayesian models~\citep{rasmussen2006}. Assume a dataset of $n$ inputs $X = (\x_1, \ldots, \x_n)$ with $\x_i \in \R^D$ and noisy outputs $\y = (y_1, \ldots, y_n) \in \R^n$. The observations $y = f(\x) + \varepsilon$ are assumed to have additive, zero mean noise $\varepsilon \sim \N(0, \sigma_y^2)$ with a zero-mean GP prior on the latent function $f(\x)$,
\begin{align}
f(\x) \sim \GP\left(0, K(\x,\x')\right),
\end{align}
which defines a distribution over functions $f(\x)$ whose mean and covariance are
\begin{align}
\E[f(\x)] &= 0 \\
\cov[ f(\x),f(\x')] &= K(\x,\x').
\end{align}
Then for any collection of inputs $X$, the function values follow a multivariate normal distribution $\f \sim \N(\0, K_{XX})$, where $\f = (f(\x_1), \ldots, f(\x_N))^T \in \R^n$, and where $K_{XX} \in \R^{n \times n}$ with $[K_{XX}]_{ij} = K(\x_i, \x_j)$. The key property of Gaussian processes is that they encode functions that predict similar output values $f(\x),f(\x')$ for similar inputs $\x,\x'$, with similarity determined by the kernel $K(\x,\x')$. In this paper we assume the Gaussian ARD kernel
\begin{align}
K(\x,\x') = \sigma_f^2 \exp\left(- \frac{1}{2} \sum_{j=1}^D \frac{ (x_j - x_j')^2}{\ell_j^2} \right),
\end{align}
with a signal variance $\sigma_f^2$ and dimension-specific lengthscale $\ell_1, \ldots, \ell_D$ parameters.

\begin{figure}[t]
\includegraphics[width=\columnwidth]{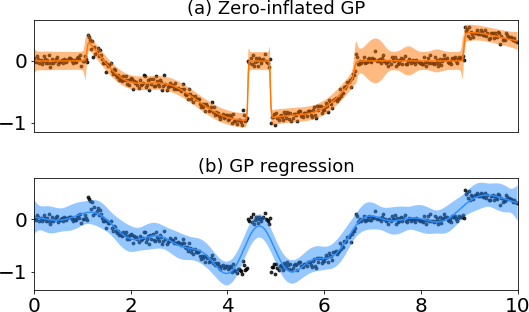}
\caption{Illustration of a zero-inflated GP \textbf{(a)} and standard GP regression \textbf{(b)}. The standard approach is unable to model sudden loss of signal (at $4 \ldots 5$) and signal close to zero (at $0 \ldots 1$ and $7 \ldots 9$).}
\label{fig:motivation}
\end{figure}

The inference of the hyperparameters $\theta = (\sigma_y, \sigma_f, \ell_1, \ldots, \ell_D)$ is performed commonly by maximizing the marginal likelihood
\begin{align}
    p(\y | \theta) &= \int p(\y | \f) p(\f | \theta) d\f,
\end{align}
which results in a convenient marginal likelihood called evidence, $p(\y | \theta) = N(\y | \0, K_{XX} + \sigma_y^2 I)$ for a Gaussian likelihood.

The Gaussian process defines a univariate normal predictive posterior distribution $f(\x) | \y,X \sim \N(\mu(\x), \sigma^2(\x))$ for an arbitrary input $\x$ with the prediction mean and variance\footnote{In the following we omit the implicit conditioning on data inputs $X$ for clarity.}
\begin{align}
\mu(\x)   &= K_{\x X} (K_{XX} + \sigma_y^2I)^{-1} \y,\\
\sigma^2(\x) &= K_{\x \x} - K_{\x X} (K_{XX} + \sigma_y^2I)^{-1}K_{X \x},
\end{align}
where $K_{X \x} = K_{\x X}^T \in \R^n$ is the kernel column vector over pairs $X \times \x$, and $K_{\x \x} = K(\x,\x) \in \R$ is a scalar. The predictions $\mu(\x) \pm \sigma(\x)$ come with uncertainty estimates in GP regression. 

\section{ZERO-INFLATED GAUSSIAN PROCESSES}

\begin{figure}[th]
\includegraphics[width=\columnwidth]{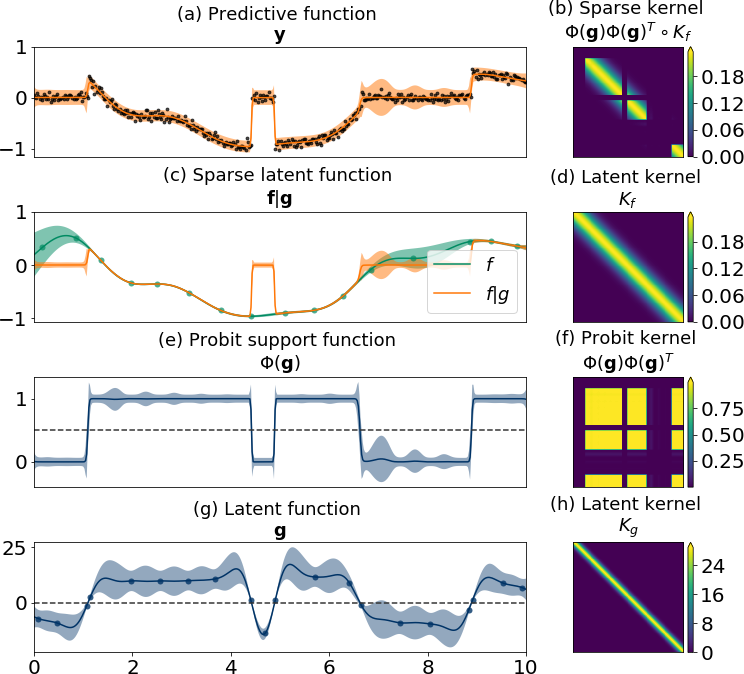}
\caption{Illustration of the zero-inflated GP \textbf{(a)} and the sparse kernel \textbf{(b)} composed of a smooth latent function \textbf{(c,d)} filtered by a probit support function \textbf{(e,f)}, which is induced by the underlying latent sparsity \textbf{(g,h)}.}
\label{fig:zigp}
\end{figure}

We introduce zero-inflated Gaussian processes that have -- in contrast to standard GP's -- a tendency to produce exactly zero predictions (See Figure \ref{fig:motivation}). Let $g(\x)$ denote the latent ``on-off" state of a function $f(\x)$. We assume GP priors for both functions with a joint model
\begin{align}
    p(\y,\f,\bg) = p(\y|\f)p(\f|\bg)p(\bg),
\end{align}
where
\begin{align}
p(\y | \f) &= \N(\y | \f, \sigma_y^2 I) \\
p(\f | \bg) &= \N(\f | \0, \Phi(\bg) \Phi(\bg)^T \circ K_f) \label{eq:sparsekernel} \\
p(\bg ) &= \N(\bg | \beta \mathbf{1}, K_g).
\end{align}
The sparsity values $g(\x)$ are squashed between $0$ and $1$ through a standard Normal cumulative distribution, or a probit link function,  $\Phi : \R \rightarrow [0,1]$
\begin{align}
\Phi(g) &= \int_{-\infty}^g \phi(\tau) d\tau = \frac{1}{2} \left(1 + \mathrm{erf}\left( \frac{g}{ \sqrt{2}}\right) \right),
\end{align}
where $\phi(\tau) = \frac{1}{\sqrt{2\pi}}e^{-\frac{1}{2}\tau^2}$ is the standard normal density function. The structured probit sparsity $\Phi(\bg)$ models the ``on-off" smoothly due to the latent sparsity function $\bg$ having a GP prior with prior mean $\beta$. The latent function $\f$ is modeled throughout but it is only visible during the ``on" states. This masking effect has similarities to both zero-inflated and hurdle models. The underlying latent function $\f$ is learned from only non-zero data similarly to in hurdle models, but the function $\f$ is allowed to predict zeros similarly to zero-inflated models.

The key part of our model is the sparse \emph{probit-sparsified covariance} $\Phi(\bg) \Phi(\bg)^T \circ K$ where the ``on-off" state $\Phi(\bg)$ has the ability to zero out rows and columns of the kernel matrix at the ``off" states (See Figure \ref{fig:zigp}f for the probit pattern $\Phi(\bg)\Phi(\bg)^T$ and Figure \ref{fig:zigp}b for the resulting sparse kernel). As the sparsity $g(\x)$ converges towards minus infinity, the probit link $\Phi(g(\x))$ approaches zero, which leads the function distribution approaching $\N(f_i | 0, 0)$, or $f_i = 0$. Numerical problems are avoided since in practice $\Phi(g) > 0$, and due to the conditioning noise variance term $\sigma_y^2 > 0$.


The marginal likelihood of the zero-inflated Gaussian process is intractable due to the probit-sparsification of the kernel. We derive a stochastic variational Bayes approximation, which we show to be tractable due to the choice of using the probit link function.

\subsection{STOCHASTIC VARIATIONAL INFERENCE}

Inference for standard Gaussian process models is difficult to scale as complexity grows with $\mathcal{O}(n^3)$ as a function of the data size $n$. \citet{titsias2009} proposed a variational inference approach for GPs using $m < n$ inducing variables, with a reduced computational complexity of $\mathcal{O}(m^3)$ for $m$ inducing points. The novelty of this approach lies in the idea that the locations and values of inducing points can be treated as variational parameters, and optimized. \citet{hensman2013,hensman2015} introduced more efficient stochastic variational inference (SVI) with factorised likelihoods that has been demonstrated with up to billion data points \citep{salimbeni2017}. This approach cannot be directly applied to sparse kernels due to having to compute expectation of the probit product in the covariance. We derive the SVI bound tractably for the zero-inflated model and its sparse kernel, which is necessary in order to apply the efficient parameter estimation techniques with automatic differentiation with frameworks such as TensorFlow \citep{abadi2016}.




We begin by applying the inducing point augmentations $f(\z_f) = \uf$ and $g(\z_g) = \ug$ for both the latent function $f(\cdot)$ and the sparsity function $g(\cdot)$. We place $m$ inducing points $\u_{f1}, \ldots \u_{fm}$ and $\u_{g1}, \ldots \u_{gm}$ for the two functions. The augmented joint distribution is $p(\y,\f,\bg,\uf,\ug) = p(\y|\f)p(\f|\bg,\uf)p(\bg|\ug)p(\uf)p(\ug)$, where\footnote{We drop the implicit conditioning on $\z$'s for clarity.}
\begin{align}
p(\f | \bg, \uf) &= \N(\f | \diag(\pbg) Q_f \uf , \pbg \pbg^T \circ \widetilde{K}_f) \\
p(\bg | \ug) &= \N(\bg | Q_g \ug , \widetilde{K}_g) \\
p(\uf) &= \N(\uf| \0, K_{f mm}) \\
p(\ug) &= \N(\ug| \0, K_{g mm})
\end{align}
and where
\begin{align}
Q_f &= K_{f nm} K_{f mm}^{-1} \label{eq:Qf} \\
Q_g &= K_{g nm} K_{g mm}^{-1} \label{eq:Qg} \\
\widetilde{K}_f &= K_{fnn} - K_{fnm}K_{fmm}^{-1}K_{fmn} \label{eq:Kf} \\
\widetilde{K}_g &= K_{g nn} -  K_{g nm}K_{g mm}^{-1}K_{g mn}. \label{eq:Kg}
\end{align}
We denote the kernels for functions $f$ and $g$ by the corresponding subscripts. The kernel $K_{fnn}$ is between all $n$ data points, the kernel $K_{fnm}$ is between all $n$ datapoints and $m$ inducing points, and the kernel $K_{fmm}$ is between all $m$ inducing points (similarly for $g$ as well).


Next we use the standard variational approach by introducing approximative variational distributions for the inducing points,
\begin{align}
q(\uf) &= \N(\uf| \mf, \Sf) \\
q(\ug) &= \N(\ug| \mg, \Sg)
\end{align}
where $\Sf,\Sg \in \R^{m \times m}$ are square positive semi-definite matrices. The variational joint posterior is
\begin{align}
q(\f,\bg,\uf,\ug) &= p(\f|\bg,\uf) p(\bg|\ug) q(\uf) q(\ug).
\end{align}
We minimize the Kullback-Leibler divergence between the true augmented posterior $p(\y,\f,\bg,\uf,\ug)$ and the variational distribution $q(\f,\bg,\uf,\ug)$, which is equivalent to solving the following evidence lower bound (as shown by e.g.~\citet{hensman2015}):
\begin{align}
    \log p(\y ) &\ge \E_{q(\f)} \log p(\y | \f) - \KL[ q(\uf,\ug) || p(\uf,\ug)], \label{eq:elbo1}
\end{align}
where we define
\begin{align}
    q(\f) &= \iiint p(\f | \bg, \uf) q(\uf) p(\bg | \ug) q(\ug) d\uf d\ug d\bg \notag \\
    &= \int q(\f | \bg) q(\bg) d\bg, \label{eq:q}
\end{align}
where the variational approximations are tractably
\begin{align}
 q(\bg) &= \int p(\bg | \ug) q(\ug) d\ug  \label{eq:varmarg} \\
  &= \N(\bg | \bmu_g, \Sigma_g) \notag \\
 q(\f | \bg) &= \int p(\f|\bg, \uf) q(\uf) d\uf \label{eq:varmarg2} \\
 &= \N(\f | \diag (\pbg) \bmu_f, \pbg \pbg^T \circ \Sigma_f) \notag
\end{align}
with
\begin{align}
\bmu_f &= Q_f \mf \label{eq:mf} \\
\bmu_g &= Q_g \mg \label{eq:mg} \\
\Sigma_f &= K_{fnn} + Q_f (\Sf - K_{fmm}) Q_f^T \label{eq:Sf} \\
\Sigma_g &= K_{gnn} + Q_g (\Sg - K_{gmm}) Q_g^T. \label{eq:Sg}
\end{align}
We additionally assume the likelihood $p(\y | \f) = \prod_{i=1}^N p(y_i | f_i)$ factorises.
We solve the final ELBO of equations \eqref{eq:elbo1} and \eqref{eq:q} as (See Supplements for detailed derivation)
\begin{align}
    \fl_{\textsc{ZI}}  &= \sum_{i = 1}^{N} \Big\{ \log \N(y_i | \langle \Phi(\g_i) \rangle_{q(\g_i)} \mu_{fi}, \Sn) \label{eq:elbozi} \\
    &\qquad - \frac{1}{2\Sn} \left ( \Var[\Phi(\g_i)] \mu^2_{fi}  + \langle \Phi(\g_i)^2 \rangle_{q(\g_i)} \sigma_{fi}^2 \right ) \Big\} \notag \\
        &\qquad - \KL[ q(\uf) || p(\uf)] - \KL[ q(\ug) || p(\ug)], \notag
\end{align}
where $\mu_{fi}$ is the $i$'th element of $\bmu_f$ and $\sigma_{fi}^2$ is the $i$'th diagonal element of $\Sigma_f$ (similarly with $g$).
The expectations are tractable,
\begin{align}
    \langle \Phi(\g_i) \rangle_{q(\g_i)} &= \Phi(\lambda_{gi} ), \qquad \lambda_{gi} = \frac{\mu_{\g i}}{\sqrt{1+\sigma_{\g i}^2}} \label{eq:T1} \\
    \langle \Phi(\g_i)^2 \rangle_{q(\g_i)} &= \Phi(\lambda_{gi}) - 2T \left( \lambda_{gi}, \frac{\lambda_{gi}}{\mu_{gi}} \right) \label{eq:T2}\\
    \Var[\Phi(\g_i)] &= \Phi(\lambda_{gi}) - 2T \left( \lambda_{gi}, \frac{\lambda_{gi}}{\mu_{gi}} \right)- \Phi(\lambda_{gi})^2. \label{eq:T3}
\end{align}
The Owen's T function $T(a,b) = \phi(a) \int_0^b \frac{\phi(a \tau)}{1+\tau^2} d\tau$ \citep{owen1956} has efficient numerical solutions in practise \citep{patefield2000}.

The ELBO is considerably more complex than the standard stochastic variational bound of a Gaussian process \citep{hensman2013}, due to the probit-sparsified covariance. 
The bound is likely only tractable for the choice of probit link function $\Phi(\bg)$, while other link functions such as the logit would lead to intractable bounds necessitating slower numerical integration \citep{hensman2015}.

We optimize the $\L_{\textsc{zi}}$ with stochastic gradient ascent techniques with respect to the inducing locations $\z_g, \z_f$, inducing value means $\mf,\mg$ and covariances $\Sf,\Sg$, the sparsity prior mean $\beta$, the noise variance $\sigma_y^2$, the signal variances $\sigma_f,\sigma_g$, and finally the dimensions-specific lengthscales $\ell_{f1}, \ldots, \ell_{fD}; \ell_{g1}, \ldots, \ell_{gD}$ of the Gaussian ARD kernel.

\section{GAUSSIAN PROCESS NETWORK}

The Gaussian Process Regression Networks (GPRN) framework by \citet{wilson2012} is an efficient model for multi-target regression problems, where each individual output is a linear but non-stationary combination of shared latent functions. Formally, a vector-valued output function $\y(\x) \in \R^P$ with $P$ outputs is modeled using vector-valued latent functions $\f(\x) \in \R^Q$ with $Q$ latent values and mixing weights $W(\x) \in \R^{P \times Q}$ as
\begin{align}
\y(x) = W(x)[\f(x) + \boldsymbol{\epsilon}] + \boldsymbol\varepsilon,
\end{align}
where for all $q = 1,\ldots,Q$ and $p = 1,\ldots,P$ we assume GP priors and additive zero-mean noises,
\begin{align}
f_q(\x) &\sim \GP(0,K_f(\x,\x')) \\
W_{qp}(\x) &\sim \GP(0,K_w(\x,\x')) \\
\epsilon_q &\sim \N(0,\sigma_f^2) \\
\varepsilon_p &\sim \N(0,\sigma_y^2).
\end{align}
The subscripts are used to denote individual components of $\f$ and $W$ with $p$ and $q$ indicating $p^{th}$ output dimension and $q^{th}$ latent dimension, respectively. We assume shared latent  and output noise variances $\sigma_f^2,\sigma_y^2$ without loss of generality. The distributions of both functions $\f$ and $W$ have been inferred either with variational EM \citep{wilson2012} or by variational mean-field approximation with diagonalized latent and mixing functions \citep{nguyen13}.

\subsection{STOCHASTIC VARIATIONAL INFERENCE}


Here, we first extend the works of \citet{wilson2012} and \citet{nguyen13} by introducing the currently missing SVI bounds for the standard GPRN, and then propose the novel sparse GPRN model, and solve its SVI bounds as well, in the following section.

We begin by introducing the inducing variable augmentation technique for latent functions $\f(\x)$ and mixing weights $W(\x)$ with $\uf,\zf = \{\u_{f_q},\z_{f_q}\}_{q=1}^Q$ and $\uw,\zw = \{\u_{w_{qp}},\z_{w_{qp}}\}_{q,p=1}^{Q,P}$:
\begin{align}
& \hspace{-13mm} p(\y, \f, W, \u_f, \u_w)  \\
&= p(\y|\f,W) p(\f | \u_f) p(W | \u_w) p(\u_f) p(\u_w) \notag \\
 p(\f | \uf) &= \prod_{q=1}^Q \N(\f_q | Q_{f_q} \u_{f_q}, \widetilde{K}_{f_q}) \\
 p(W | \uw) &= \prod_{q,p=1}^{Q,P} \N(\w_{qp} | Q_{w_{qp}} \u_{w_{qp}} , \widetilde{K}_{w_{qp}}) \\
 p(\uf) &= \prod_{q=1}^Q \N(\u_{f_q}| \0, K_{f_q,mm}) \\
 p(\uw) &= \prod_{q,p=1}^{Q,P} \N(\u_{w_{qp}} | \0, K_{w_{qp},mm}),
\end{align}
where we have separate kernels $K$ and extrapolation matrices $Q$ for each component of $W(\x)$ and $\f(\x)$ that are of the same form as in \cref{eq:Qf,eq:Qg,eq:Kf,eq:Kg}. The $\w$ is a vectorised form of $W$. The variational approximation is then 
\begin{align}
q(\f,W,\u_f,\u_w) &= p(\f | \u_f) p(W | \u_w)  q(\u_f) q(\u_w) \\
q(\u_{f_q}) &= \prod_{q=1}^Q \N(\u_{f_q} | \m_{f_q}, \mathbf{S}_{f_q}) \\
q(\u_{w_{qp}}) &= \prod_{q,p=1}^{Q,P} \N(\u_{w_{qp}} | \m_{w_{qp}}, \mathbf{S}_{w_{qp}}),
\end{align}
where $\u_{w_{qp}}$ and $\u_{f_q}$ indicate the inducing points for the functions $W_{qp}(\x)$ and $f_q(\x)$, respectively. The ELBO can be now stated as
\begin{align}
    \log p(\y ) &\ge \E_{q(\f,W)} \log p(\y | \f, W) \label{eq:elbo2} \\
     &\qquad - \KL[ q(\u_f, \u_w) || p(\u_f,\u_w)], \notag
\end{align}
where the variational distributions decompose as $q(\f,W) = q(\f) q(W)$ with marginals of the same form as in \cref{eq:mf,eq:mg,eq:Sf,eq:Sg},
\begin{align}
q(\f) &= \int \hspace{-0.5mm} q(\f | \u_f) q(\u_f) d\u_f = \N(\f | \bmu_f,\S_f) \\
q(W)  &= \int \hspace{-0.5mm} q(W | \u_w) q(\u_w) d \u_w = \N(\w | \bmu_w,\S_w).
\end{align}


Since the noise term $\boldsymbol\varepsilon$ is assumed to be isotropic Gaussian, the density $p(\y|W,\f)$ factorises across all target observations and dimensions. The expectation term in equation (\ref{eq:elbo2}) then reduces to solving the following integral for the $i^{th}$ observation and $p^{th}$ target dimension,
\begin{align}
  \sum_{i,p=1}^{N,P} \iint \log \N(y_{p,i} |\w_{p,i}^T \f_i, \sigma_y^2) q(\f_i,\w_{p,i}) d\w_{p,i} d\f_i .
\end{align}
The above integral has a closed form solution resulting in the final ELBO as (See Supplements)
\begin{align}
    \fl_{\textsc{gprn}}  &= \sum_{i=1}^{N} \Bigg\{ \sum_{p=1}^{P}  \log \N \Big(y_{p,i} | \sum_{q=1}^Q  \mu_{w_{qp},i} \mu_{f_q,i}, \sigma_{y}^2 \Big) \notag \\
     &\hspace{-9mm} - \frac{1}{2\sigma_{y}^2} \sum_{q,p=1}^{Q,P} \left ( \mu_{w_{qp},i}^2 \sigma^2_{f_q,i} + \mu_{f_q,i}^2 \sigma_{w_{qp},i}^2 + \sigma_{w_{qp},i}^2 \sigma_{f_q,i}^2 \right) \hspace{-1.5mm} \Bigg\} \notag \\
     &\hspace{-9mm} - \sum_{q,p}^{Q,P} \KL[ q(\u_{w_{qp}},\u_{f_q}) || p(\u_{w_{qp}},\u_{f_q})],
\end{align}
where $\mu_{f_{q},i}$ is the $i$'th element of $\bmu_{f_q}$ and $\sigma_{f_{q},i}^2$ is the $i$'th diagonal element of $\Sigma_{f_q}$ (similarly for the $W_{qp}$'s).

\section{SPARSE GAUSSIAN PROCESS NETWORK}

In this section we demonstrate how zero-inflated GPs can be used as plug-in components in other standard models. In particular, we propose a significant modification to GPRN by adding sparsity to the mixing matrix components. This corresponds to each of the $p$ outputs being a sparse mixture of the latent $Q$ functions, i.e. they can effectively use any subset of the $Q$ latent dimensions by having zeros for the rest in the mixing functions. This makes the mixture more easily interpretable, and induces a variable number of latent functions to explain the output of each input $\x$. The latent function $\f$ can also be sparsified, with a derivation analogous to the derivation below.

We extend the GPRN with probit sparsity for the mixing matrix $W$, resulting in a joint model
\begin{align}
p(\y, \f, W, \bg) &= p(\y|\f,W) p(\f) p(W | \bg) p(\bg),
\end{align}
where all individual components of the latent function $\f$ and mixing matrix $W$ are given GP priors. We encode the sparsity terms $\bg$ for all the $Q \times P$ mixing functions $W_{qp}(\x)$ as
\begin{align}
p(W_{qp}| \bg_{qp}) &= \N(\w_{qp}| \0, \Phi(\bg_{qp}) \Phi(\bg_{qp})^T \circ K_w ).
\end{align}
To introduce variational inference, the joint model is augmented with three sets of inducing variables for $\f$, $W$ and $\bg$. After marginalizing out the inducing variables as in \cref{eq:q,eq:varmarg,eq:varmarg2}, the marginal likelihood can be written as
\begin{align}
    \log p(\y) &\ge \E_{q(\f,W,
    \bg)} \log p(\y | \f, W) \label{eq:elbo3} \\
     &\qquad - \KL[ q(\u_f,\u_w,\u_g) || p(\u_f,\u_w,\u_g)]. \notag
\end{align}
The joint distribution in the variational expectation factorizes as $q(\f,W,\bg) = q(\f) q(W|\bg) q(\bg)$. Also, with a Gaussian noise assumption, the expectation term factories across all the observations and target dimensions. The key step reduces to solving the following integrals:
\begin{align}
\sum_{i,p=1}^{N,P} \iiint \log \N(y_{p,i} |(\w_{p,i} \circ \bg_{p,i})^T \f_i, \sigma_y^2) \\ \cdot \, q(\f_i,\w_{p,i},\bg_{p,i}) d\w_{p,i} d\f_i d\bg_{p,i}. \notag
\end{align}
The above integral has a tractable solution leading to the final sparse GPRN evidence lower bound (See Supplements)
\begin{align}
    \fl_{s\textsc{gprn}}  &= \sum_{i=1}^{N} \Bigg\{ \sum_{p=1}^{P}  \log \N \Big(y_{p,i} | \sum_{q=1}^Q  \mu_{w_{qp},i} \mu_{g_{qp},i} \mu_{f_q,i}, \sigma_y^2 \Big) \notag \\
     & \hspace{-8mm} - \frac{1}{2\sigma_y^2} \sum_{q,p=1}^{Q,P} \Big((\mu_{g_{qp},i}^2 + \sigma_{g_{qp},i}^2)  \\
     & \hspace{8mm} \cdot (\mu_{w_{qp},i}^2  \sigma_{f_q,i}^2  + \mu_{f_q,i}^2 \sigma_{w_{qp},i}^2 + \sigma_{w_{qp},i}^2 \sigma_{f_q,i}^2 ) \Big) \notag \\
     & \hspace{-8mm} - \frac{1}{2\sigma_y^2} \sum_{q,p=1}^{Q,P} \left (\sigma_{g_{qp},i}^2  \mu_{f_q,i}^2 \mu_{w_{qp},i}^2\right) \Bigg\} \notag \\
     & \hspace{-8mm} - \sum_{q,p}^{Q,P} \KL[ q(\u_{f_{q}},\u_{w_{qp}},\u_{g_{qp}}) || p(\u_{f_q},\u_{w_{qp}},\u_{g_{qp}})], \notag
\end{align}
where $\mu_{f_q,i}, \mu_{w_{qp},i}$ are the variational expectation means for $\f(\cdot), W(\cdot)$ as in  \cref{eq:mf,eq:mg}, $\mu_{g_{qp},i}$ is the variational expectation mean of $g(\cdot)$ as in equation \eqref{eq:T1}, and analogously for the variances.


\section{EXPERIMENTS}

First we demonstrate how the proposed method can be used for regression problems with zero-inflated targets. We do that both on a simulated dataset and for real-world climate modeling scenarios on a Finnish rain precipitation dataset with approximately 90\% zeros. Finally, we demonstrate the GPRN model and how it improves both the interpretability and predictive performance in the JURA geological dataset.

We use the squared exponential kernel with ARD in all experiments. All the parameters including inducing locations, values and variances and kernel parameters were learned through stochastic Adam optimization \citep{kingma2014} on the TensorFlow \citep{abadi2016} platform.

We compare our approach \textsc{ZiGP} to baseline \textsc{Zero} voting, to conventional Gaussian process regression (\textsc{GPr}) and classification (\textsc{GPc}) with SVI approximations from the GPflow package \citep{GPflow2017}. Finally, we also compare to first classifying the non-zeros, and successively applying regression either to all data points (\textsc{GPcr}), or to only predicted non-zeros (\textsc{GPcr$_{\not = 0}$}, hurdle model).

We record the predictive performance by considering mean squared error and mean absolute error. We also compare the models' ability to predict true zeros with F1, accuracy, precision, and recall of the optimal models. 

\begin{figure}[t]
\includegraphics[width=\columnwidth]{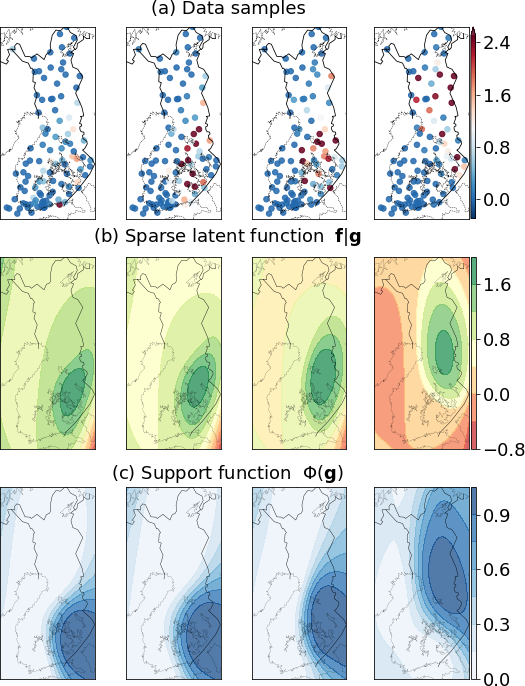}
\caption{\textsc{ZiGP} model fit on the precipitation dataset. Sample of the actual data \textbf{(a)} against the sparse rain function estimate \textbf{(b)}, with the probit support function \textbf{(c)} showing the rain progress.}
\label{fig:pptr1}
\end{figure}

\begin{table}[t]
\caption{Results for the precipitation dataset over baseline (Zero; majority voting), four competing methods and the proposed method \textsc{ZiGP} on test data. The columns list both quantitative and qualitative performance criteria, best performance is boldfaced.}
\label{tab:rain}
\vskip 0.15in
\begin{center}
\begin{sc}
\begin{small}
\resizebox{\columnwidth}{!}{
\begin{tabular}{l|cc|ccccc}
\hline
Model  &  RMSE & MAE & F1 & Acc. & Prec. & Recall \\
\hline
Zero & 0.615 & 0.104 & 0.000 & 0.898 & 0.000 & 0.000 \\
GPc      & -     & -     & 0.367 & \textbf{0.911} & 0.675 & 0.252 \\
GPr      & 0.569 & 0.159 & 0.401 & 0.750 &0.266 & \textbf{0.817} \\
GPcr     & 0.589 & \textbf{0.102} & 0.366 & \textbf{0.911} & 0.679 & 0.251 \\
GPcr$_{\not = 0}$ & 0.575 & \textbf{0.101} & 0.358 & \textbf{0.912} & \textbf{0.712} & 0.240 \\
ZiGP     & \textbf{0.561} & 0.121 & \textbf{0.448} & 0.861 & 0.381 & 0.558 \\
\hline
\end{tabular}
}
\end{small}
\end{sc}
\end{center}
\vskip -0.1in
\end{table}





\subsection{SPATIO-TEMPORAL DATASET}

Zero-inflated cases are commonly found in climatology and ecology domains. In this experiment we demonstrate the proposed method by modeling precipitation in Finland\footnote{Data can be found at \url{http://en.ilmatieteenlaitos.fi/}}. The dataset consists of hourly quantitative non-negative observations of precipitation amount across 105 observatory locations in Finland for the month of July 2018. The dataset contains 113015 datapoints with approximately 90\% zero precipitation observations. The data inputs are three-dimensional: latitude, longitude and time. Due to the size of the data, this experiment illustrates the scalability of the variational inference.

We randomly split 80\% of the data for training and the rest 20\% for testing purposes. We split across time only, such that at a single measurement time,  all locations are simultaneously either in the training set, or in the test set.

We further utilize the underlying spatio-temporal grid structure of the data to perform inference in an efficient manner by Kronecker techniques \citep{saatchi2011}. All the kernels for latent processes are assumed to factorise as $\K = \K_{space} \otimes \K_{time}$ which allows placing inducing points independently on spatial and temporal grids.

\begin{figure}[ht]
\includegraphics[width=\columnwidth]{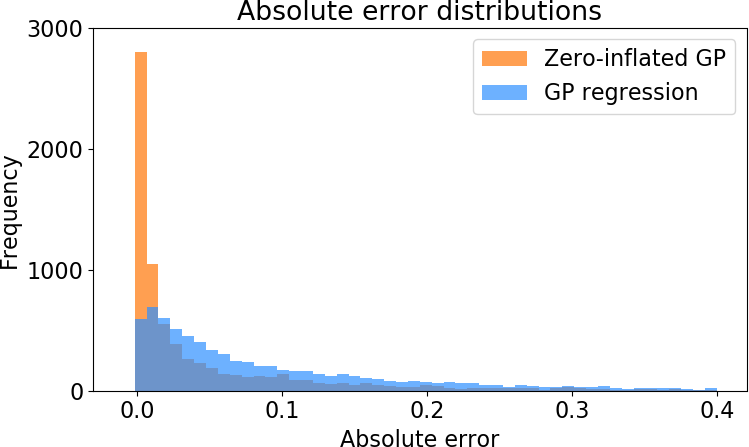}
\caption{The distribution of errors with the rain dataset with the \textsc{ZiGP} and the \textsc{GPr}. The zero-inflated GP achieves much higher number of perfect (zero) predictions.}
\label{fig:pptr2}
\end{figure}

Figure \ref{fig:pptr1} depicts the components of the zero-inflated GP model on the precipitation dataset. As shown in panel (c), the latent support function models the presence or absence of rainfall. It smoothly follows the change in rain patterns across hourly observations. The amount of precipitation is modeled by the other latent process and the combination of these two results in sparse predictions. Figure \ref{fig:pptr2} shows that the absolute error distribution is remarkably better with the \textsc{ZiGP} model due to it identifying the absence of rain exactly. While both models fit the high rainfall regions well, for zero and near-zero regions \textsc{GPr} does not refine its small errors. Table \ref{tab:rain} indicates that the \textsc{ZiGP} model achieves the lowest mean square error, while also achieving the highest F1 score that takes into account the class imbalance, which biases the elementary accuracy, precision and recall quantities towards the majority class.


\subsection{MULTI-OUTPUT PREDICTION}

In this experiment we model the multi-response Jura dataset with the sparse Gaussian process regression network \textsc{sGPRN} model and compare it with standard \textsc{GPRN} as baseline. Jura contains concentration measurements of cadmium, nickel and zinc metals in the region of Swiss Jura. We follow the experimental procedure of \citet{wilson2012} and \citet{nguyen13}. The training set consists of $n=259$ observations across $D=2$ dimensional geo-spatial locations, and the test set consists of 100 separate locations. For both models we use $Q=2$ latent functions with the stochastic variational inference techniques proposed in this paper. Sparse GPRN uses a sparsity inducing kernel in the mixing weights. The locations of inducing points for the weights $W(\x)$ and the support $g(\x)$ are shared. The kernel length-scales are given a gamma prior with the shape parameter $\alpha = 0.3$ and rate parameter $\beta = 1.0$ to induce smoothness. We train both the models 30 times with random initialization.

\begin{figure}[t]
\includegraphics[width=\columnwidth]{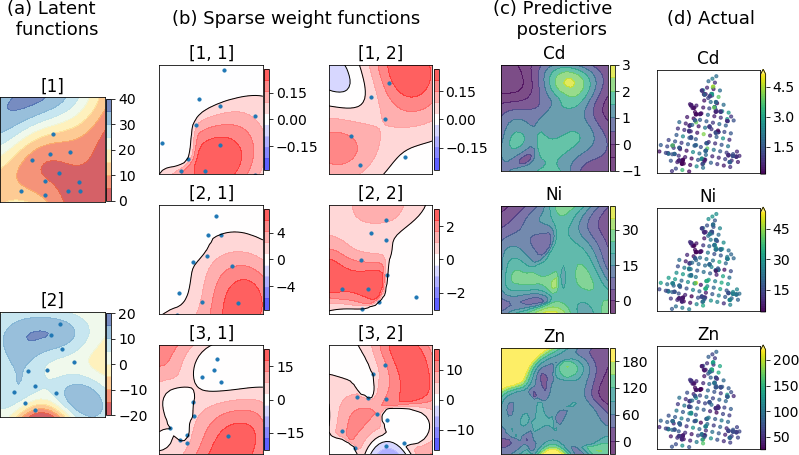}
\caption{The sparse GPRN model fit on the Jura dataset with 11 inducing points. The $Q=2$ (dense) latent functions \textbf{(a)} are combined with the $3 \times 2$ sparse mixing functions \textbf{(b)} into the $P=3$ output predictions \textbf{(c)}. The real data are shown in \textbf{(d)}. The white mixing regions are estimated `off'.}
\label{fig:gprn1}
\end{figure}

\begin{figure}[h]
\includegraphics[width=\columnwidth]{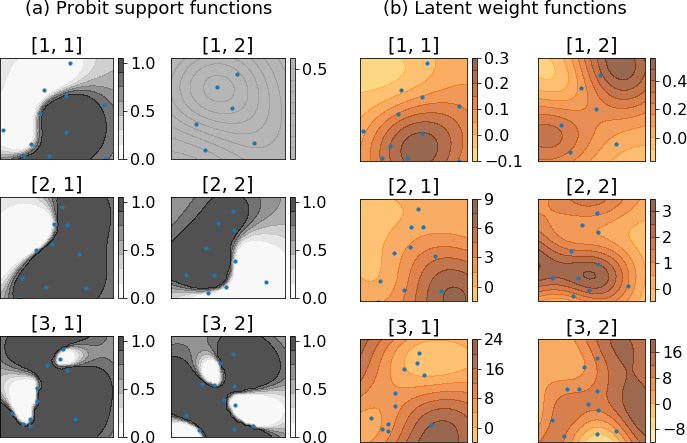}
\caption{The sparse probit support \textbf{(a)} and latent functions \textbf{(b)} of the weight function $W(\x)$ of the optimized sparse GPRN model. The black regions of \textbf{(a)} show regional activations, while the white regions show where the latent functions are `off'. The elementwise product of the support and weight functions is indicated in the Figure \ref{fig:gprn1}\textbf{b)}.}
\label{fig:gprn2}
\end{figure}

Table \ref{tab:gprn} shows that our model performs better than the state-of-the-art SVI-GPRN, both with $m=5$ and $m=10$ inducing points. Figure \ref{fig:gprn1} visualises the optimized sparse GPRN model, while Figure \ref{fig:gprn2} indicates the sparsity pattern in the mixing weights. The weights have considerable smooth `on' regions (black), and also interesting smooth `off' regions (white). The `off' regions indicate that for certain locations, only one of the two latent functions is adaptively utilised.

\begin{table}[t]
\label{tab:gprn}
\caption{Results for the Jura dataset for sparse GPRN and vanilla GPRN models with test data. Best performance is with boldface. We do not report RMSE and MAE values \textsc{GPc}, since its a classification method.}
\vskip 0.15in
\begin{center}
\begin{sc}
\begin{small}
\resizebox{\columnwidth}{!}{
\begin{tabular}{lc|cc|cc|cc}
\hline
Model & $m$ & RMSE & MAE &  RMSE & MAE & RMSE & MAE\\
\hline
\textsc{GPRN}  & 5    &        0.732    &  0.572    &   6.807    &  5.163    &   \textbf{34.41}   &  \textbf{22.14}    \\
\textsc{sGPRN} & 5    &        \textbf{0.728}    &  \textbf{0.567}    &   \textbf{6.631}    &  \textbf{5.079}    &   35.09   &  22.75    \\
\textsc{GPRN}  & 10   &        0.774    &  0.586    &   7.207    &  5.656    &   37.87   &  25.10    \\
\textsc{sGPRN} & 10  &        \textbf{0.749}    &  \textbf{0.573}    &   \textbf{6.524}    &  \textbf{5.054}    &   \textbf{36.17}   &  \textbf{23.63}    \\
\hline
\end{tabular}
}
\end{small}
\end{sc}
\end{center}
\vskip -0.1in
\end{table}

\section{DISCUSSION}

We proposed a novel paradigm of zero-inflated Gaussian processes with a novel sparse kernel. The sparsity in the kernel is modeled with smooth probit filtering of the covariance rows and columns. This model induces zeros in the prediction function outputs, which is highly useful for zero-inflated datasets with excess of zero observations. Furthermore, we showed how the zero-inflated GP can be used to model sparse mixtures of latent signals with the proposed sparse Gaussian process network. The latent mixture model with sparse mixing coefficients leads to locally using only a subset of the latent functions, which improves interpretability and reduces model complexity. We demonstrated tractable solutions to stochastic variational inference of the sparse probit kernel for the zero-inflated GP, conventional GPRN, and sparse GPRN models, which lends to efficient exploration of the parameter space of the model. 







\bibliography{refs}
\bibliographystyle{plainnat}

\clearpage
\onecolumn
\renewcommand{\theequation}{S.\arabic{equation}}
\setcounter{equation}{0}

\section*{Supplementary material for the ``Zero-inflated Gaussian processes with sparse kernels''}

Here we show in detail how we arrived at the three new evidence lower bounds for the zero-inflated Gaussian process, for the Gaussian process network, and for the sparse Gaussian process network.

\subsection*{A) The stochastic variational bound of the zero-inflated GP}

Here, we will derive the evidence lower bound (ELBO) of the zero-inflated Gaussian process. We show how to solve the ELBO of equations \eqref{eq:elbo1} and \eqref{eq:q}, which results in the equation \eqref{eq:elbozi}.

The augmented true model with inducing points is defined
\begin{align}
p(\y,\f,\bg,\uf,\ug) &= p(\y|\f)p(\f|\bg,\uf)p(\bg|\ug)p(\uf)p(\ug) \\
p(\y | \f) &= \N(\y | \f, \Sn I) \\
p(\f | \bg, \uf) &= \N(\f | \diag(\pbg) K_{f nm} K_{f mm}^{-1} \uf , \pbg \pbg^T \circ \widetilde{K}_f) \\
p(\bg | \ug) &= \N(\bg | K_{g nm} K_{g mm}^{-1} \ug , \widetilde{K}_g) \\
p(\uf) &= \N(\uf| \0, K_{f mm}) \\
p(\ug) &= \N(\ug| \0, K_{g mm}).
\end{align}
We define the variational posterior approximation as
\begin{align}
q(\f,\bg,\uf,\ug) &= p(\f|\bg,\uf) p(\bg|\ug) q(\uf) q(\ug) \\
p(\f | \bg, \uf) &= \N(\f | \diag(\pbg) K_{f nm} K_{f mm}^{-1} \uf , \pbg \pbg^T \circ \widetilde{K}_f) \\
p(\bg | \ug) &= \N(\bg | K_{g nm} K_{g mm}^{-1} \ug , \widetilde{K}_g) \\
q(\uf) &= \N(\uf| \mf, \Sf) \\
q(\ug) &= \N(\ug| \mg, \Sg)
\end{align}
and where $\Sf,\Sg \in \R^{m \times m}$ are square positive semi-definite matrices, and we define shorthands
\begin{align}
\widetilde{K}_f &= K_{fnn} - K_{fnm}K_{fmm}^{-1}K_{fmn} \\
\widetilde{K}_g &= K_{g nn} -  K_{g nm}K_{g mm}^{-1}K_{g mn}.
\end{align}

In variational inference we minimize the Kullback-Leibler divergence between the variational approximation $q(\f,\bg,\uf,\ug)$ and the true augmented joint distribution $p(\y,\f,\bg,\uf,\ug)$:
\begin{align}
    \KL[ q(\f,\bg,\uf,\ug) || p(\y,\f,\bg,\uf,\ug)] &= \int q(\f,\bg,\uf,\ug) \log \frac{p(\y,\f,\bg,\uf,\ug)}{q(\f,\bg,\uf,\ug)} d\f d\bg d\uf d\ug \\
    &= \int q(\f,\bg,\uf,\ug) \log \frac{p(\y|\f) p(\f|\bg,\uf) p(\bg|\ug)p(\uf) p(\ug)}{p(\f|\bg,\uf) p(\bg|\ug) q(\uf) q(\ug)} d\f d\bg d\uf d\ug \\
    &= \int q(\f,\bg,\uf,\ug) \log \frac{p(\y|\f) p(\uf) p(\ug)}{q(\uf) q(\ug)} d\f d\bg d\uf d\ug \\
    &= \iiiint p(\f|\bg,\uf) p(\bg|\ug) q(\uf) q(\ug) \log p(\y|\f) d\uf d\ug d\bg d\f  \\
    &\qquad - \underbrace{\int q(\uf) \log  q(\uf) d\uf}_{\KL[q(\uf) || p(\uf)]} - \underbrace{\int q(\ug) \log  q(\ug) d\ug}_{\KL[q(\ug) || p(\ug)]}
\end{align}
Following the derivation of \citet{hensman2015}, this corresponds to maximizing the evidence lower bound (ELBO) of equation \eqref{eq:elbo1}:
\begin{align}
    \log p(\y ) &\ge \iiiint \log p(\y | \f) p(\f | \bg, \uf) q(\uf) p(\bg | \ug) q(\ug) d\uf d\ug d\bg d\f - \KL[ q(\uf,\ug) || p(\uf,\ug)] \\
    &= \E_{q(\f)} \log p(\y | \f) - \KL[ q(\uf,\ug) || p(\uf,\ug)]
\end{align}
where we define
\begin{align}
    q(\f) &= \iiint p(\f | \bg, \uf) q(\uf) p(\bg | \ug) q(\ug) d\uf d\ug d\bg \notag \\
    &= \int q(\f | \bg) q(\bg) d\bg,
\end{align}
where the variational approximations are tractably
\begin{align}
 q(\bg) &= \int p(\bg | \ug) q(\ug) d\ug   \\
  &= \int \N(\bg | K_{g nm} K_{g mm}^{-1} \ug , \widetilde{K}_g) \N(\ug| \mg, \Sg) d\ug \\
  &= \N(\bg | \bmu_g, \Sigma_g)  \\
 q(\f | \bg) &= \int p(\f|\bg, \uf) q(\uf) d\uf \\
  &= \int \N(\f | \diag(\pbg) K_{f nm} K_{f mm}^{-1} \uf , \pbg \pbg^T \circ \widetilde{K}_f) \N(\uf| \mf, \Sf) d\uf \\
 &= \N(\f | \diag (\pbg) \bmu_f, \pbg \pbg^T \circ \Sigma_f)
\end{align}
with
\begin{align}
\bmu_f &= K_{fnm} K_{fmm}^{-1} \mf  \\
\bmu_g &= K_{gnm} K_{g mm}^{-1} \mg  \\
\Sigma_f &= K_{fnn} + K_{fnm} K_{fmm}^{-1} (\Sf - K_{fmm}) K_{fmm}^{-1} K_{fmn} \\
\Sigma_g &= K_{gnn} + K_{g nm} K_{g mm}^{-1} (\Sg - K_{gmm}) K_{g mm}^{-1} K_{g mn}.
\end{align}
The variational marginalizations $q(\bg)$ and $q(\f | \bf)$ follow from standard Gaussian identities\footnote{See for instance Bishop (2006): Pattern recognition and Machine learning, Springer, Section 2.3.}. Substituting the variational marginalizations back to the ELBO results in
\begin{align}
    \log p(\y ) &\ge \int \int \log p(\y | \f) q(\f | \bg) q(\bg) d\f d\bg - \KL[ q(\uf,\ug) || p(\uf,\ug)].
\end{align}

Next, we marginalize the $\f$ from the ELBO. We additionally assume the likelihood $p(\y | \f) = \prod_{i=1}^N p(y_i | f_i)$ factorises, which results in
\begin{align}
    \int_\f \log p(\y|\f) q(\f|\bg) d\f &= \int \log \N(\y | \f, \Sn I) q(\f|\bg) d\f\\
    &= \sum_{i = 1}^{N} \int \log \N(y_i | f_i, \Sn) q(f_i|g_i) d f_i\\
    &= \sum_{i = 1}^{N} \log \N(y_i | \Phi(\g_i) \bk_{fi}^T K_{fmm}^{-1} m_{fi}, \Sn) - \frac{1}{2\Sn} \left \{ \Phi(\g_i)^2(\bk_{fii} + \bk_{fi}^T K_{fmm}^{-1} (\Sf - K_{fmm}) K_{fmm}^{-1} \bk_{fi}) \right \}\\
    &= \sum_{i = 1}^{N} \log \N(y_i | \Phi(\g_i) \mu_{fi}, \Sn) - \frac{1}{2\Sn} \left \{ \Phi(\g_i)^2 \sigma_{fi}^2 \right \},
\end{align}
where
\begin{align}
\mu_{fi} &= [\bmu_f]_i = \bk_{fi}^T K_{fmm}^{-1} m_{fi}\\
\sigma_{fi}^2 &= [\Sigma_f]_{ii} = \bk_{fii}+ \bk_{fi}^T K_{fmm}^{-1} (\Sf - K_{fmm}) K_{fmm}^{-1} \bk_{fi}.
\end{align}

Substituting the above result into the ELBO results in
\begin{align}
    \E_{q(\f)} \log p(\y | \f) &= \int_\bg q(\bg) \int_\f q(\f|\bg) \log p(\y|\f)d\f d\bg\\
    &= \int_\bg \sum_{i = 1}^{N} \log \N(y_i | \Phi(\g_i)\mu_{fi}, \Sn) - \frac{1}{2\Sn} \left \{ \Phi(\g_i)^2  \sigma_{fi}^2 \right \} \: q(\bg) d\bg\\
    &= \sum_{i = 1}^{N} \int_{\g_i} \log \N(y_i | \Phi(\g_i)\mu_{fi}, \Sn) \: q(\g_i) d\g_i -  \frac{1}{2\Sn} \sum_{i = 1}^{N}  \int_{\g_i} \left \{ \Phi(\g_i)^2  \sigma_{fi}^2 \right \} \: q(\g_i) d\g_i\\
    &= \sum_{i = 1}^{N} \log \N(y_i | \langle \Phi(\g_i) \rangle_{q(\g_i)} \mu_{fi}, \Sn) - \frac{1}{2\Sn} \sum_{i = 1}^{N} \left \{ Var[\Phi(\g_i)] (\mu_{fi})^2 \right \}  -  \frac{1}{2\Sn} \sum_{i = 1}^{N} \left \{ \langle \Phi(\g_i)^2 \rangle_{q(\g_i)} \sigma_{fi}^2 \right \}.
\end{align}
The expectations $\langle . \rangle_{q(\g_i)}$ of CDF transformation of a random variable with univariate Gaussian distribution. The analytical forms for these integrals can be written:
\begin{align}
    \langle \Phi(\g_i) \rangle_{q(\g_i)} &= \int \Phi(\g_i) q(\g_i) d\g_i \\
    &= \int \Phi(\g_i) \N(\g_i | \mu_{\g i}, \sigma_{\g i}^2) d\g_i\\
    &= \Phi \left(\frac{\mu_{\g i}}{\sqrt{1+\sigma_{\g i}^2}} \right) \\
    Var[\Phi(\g_i)] &= \int (\Phi(\g_i)-\langle \Phi(\g_i) \rangle_{q(\g_i)})^2 q(\g_i) d\g_i \\
    &= \Phi \left(\frac{\mu_{\g i}}{\sqrt{1+\sigma_{\g i}^2}} \right) - 2T \left ( \frac{\mu_{\g i}}{\sqrt{1+\sigma_{\g i}^2}},\frac{1}{\sqrt{1+2 \: \sigma_{\g i}^2}} \right ) -  \Phi \left(\frac{\mu_{\g i}}{\sqrt{1+\sigma_{\g i}^2}} \right) ^2\\
    \langle \Phi(\g_i)^2 \rangle_{q(\g_i)} &= \int \Phi(\g_i)^2 q(\g_i) d\g_i \\
    &= \Phi \left(\frac{\mu_{\g i}}{\sqrt{1+\sigma_{\g i}^2}} \right) - 2T \left ( \frac{\mu_{\g i}}{\sqrt{1+\sigma_{\g i}^2}},\frac{1}{\sqrt{1+2 \: \sigma_{\g i}^2}} \right )
\end{align}
where
\begin{align}
\mu_{\g i} &= [\bmu_g]_i = \bk_{\g i}^T K_{\g mm}^{-1} m_{\g i}\\
\sigma_{\g i}^2 &= [\Sigma_g]_{ii} = K_{\g ii} + \bk_{\g i}^T K_{\g mm}^{-1} (\Sg - K_{\g mm}) K_{\g mm}^{-1} \bk_{\g i}.
\end{align}
Owen's T function is defined as $T(h,a) = \phi(h) \int_0^a \frac{\phi(hx)}{1+x^2} dx$.

The final evidence lower bound with the Kullback-Leibler terms is
\begin{align}
    p(\y)  &\ge \sum_{i = 1}^{N} \left \{ \log \N(y_i | \langle \Phi(\g_i) \rangle_{q(\g_i)} \mu_{fi}, \Sn) - \frac{1}{2\Sn} \left ( Var[\Phi(\g_i)] \: \mu^2_{fi}  + \langle \Phi(\g_i)^2 \rangle_{q(\g_i)} \: \sigma_{fi}^2 \right ) \right \}\\
        &\qquad - \left \{ \frac{1}{2} \log \left | K_{fmm} \right | - \frac{1}{2} \log \left | \Sf \right | + \frac{1}{2} Tr \left [ (\mf \mf^T + \Sf)K_{fmm}^{-1} \right] - \frac{m}{2} \right \}\\
        &\qquad - \left \{ \frac{1}{2} \log \left | K_{\g mm} \right | - \frac{1}{2} \log \left | \Sg \right | + \frac{1}{2} Tr \left [ (\mg \mg^T + \Sg)K_{\g mm}^{-1} \right] - \frac{m}{2} \right \} \\
        &= \fl_{\textsc{ZiGP}}
\end{align}

\subsection*{B) The stochastic variational bound of the Gaussian process network}
In GPRN a vector-valued output function $\y(\x) \in \R^P$ with $P$ outputs is modeled using vector-valued latent functions $\f(\x) \in \R^Q$ with $Q$ latent values and mixing weights $W(\x) \in \R^{P \times Q}$ as
\begin{align}
\y(x) = W(x)[\f(x) + \boldsymbol{\epsilon}] + \boldsymbol\varepsilon,
\end{align}
where for all $q = 1,\ldots,Q$ and $p = 1,\ldots,P$ we assume GP priors and additive zero-mean noises,
\begin{align}
f_q(\x) &\sim \GP(0,K_f(\x,\x')) \\
W_{qp}(\x) &\sim \GP(0,K_w(\x,\x')) \\
\epsilon_q &\sim \N(0,\sigma_f^2) \\
\varepsilon_p &\sim \N(0,\sigma_y^2).
\end{align}
The subscripts are used to denote individual components of $\f$ and $W$ with $p$ and $q$ indicating $p^{th}$ output dimension and $q^{th}$ latent dimension respectively.

We begin by introducing the inducing variable augmentation for latent functions $\f(\x)$ and mixing weights $W(\x)$ with $\uf,\zf = \{\u_{f_q},\z_{f_q}\}_{q=1}^Q$ and $\uw,\zw = \{\u_{w_{qp}},\z_{w_{qp}}\}_{q,p=1}^{Q,P}$:
\begin{align}
 p(\y, \f, W, \u_f, \u_w) &= p(\y|\f,W) p(\f | \u_f) p(W | \u_w) p(\u_f) p(\u_w)  \\
 p(\f | \uf) &= \prod_{q=1}^Q \N(\f_q | Q_{f_q} \u_{f_q}, \widetilde{K}_{f_q}) \\
 p(W | \uw) &= \prod_{q,p=1}^{Q,P} \N(\w_{qp} | Q_{w_{qp}} \u_{w_{qp}} , \widetilde{K}_{w_{qp}}) \\
 p(\uf) &= \prod_{q=1}^Q \N(\u_{f_q}| \0, K_{f_q,mm}) \\
 p(\uw) &= \prod_{q,p=1}^{Q,P} \N(\u_{w_{qp}} | \0, K_{w_{qp},mm}),
\end{align}
where we have separate kernels $K$ and extrapolation matrices $Q$ for each component of $W(\x)$ and $\f(\x)$ that are of the form as given below:
\begin{align}
Q_f &= K_{f_q nm} K_{f_q mm}^{-1}  \\
Q_g &= K_{w_{qp} nm} K_{w_{qp} mm}^{-1} \\
\widetilde{K}_f &= K_{f_qnn} - K_{f_qnm}K_{f_qmm}^{-1}K_{f_qmn}  \\
\widetilde{K}_{w_{qp}} &= K_{w_{qp} nn} -  K_{w_{qp} nm}K_{w_{qp} mm}^{-1}K_{w_{qp} mn}.
\end{align}

Following the variational inference framework, we define the variational joint distribution as,
\begin{align}
q(\f,W,\u_f,\u_w) &= p(\f | \u_f) p(W | \u_w)  q(\u_f) q(\u_w) \label{eq:Qjoint} \\
q(\u_{f_q}) &= \prod_{q=1}^Q \N(\u_{f_q} | \m_{f_q} \mathbf{S}_{f_q}) \\
q(\u_{w_{qp}}) &= \prod_{q,p=1}^{Q,P} \N(\u_{w_{qp}} | \m_{w_{qp}}, \mathbf{S}_{w_{qp}}),
\end{align}
where $\u_{w_{qp}}$ and $\u_{f_q}$ indicate the inducing points for functions $W_{qp}(\x)$ and $f_q(\x)$, respectively. The ELBO can be now stated as
\begin{align}
    \log p(\y ) &\ge \E_{q(\f,W,\u_f,\u_w)} \log p(\y | \f, W)  - \KL[ q(\u_f, \u_w) || p(\u_f,\u_w)] \\
     &= \iiiint q(\f,W,\u_f,\u_w) \log p(\y | \f, W) d\f dW d\u_f d\u_w - \KL[ q(\u_f, \u_w) || p(\u_f,\u_w)]
\end{align}
Since the variational joint posterior decomposes as \cref{eq:Qjoint}, we begin by marginalizing the inducing distributions $\u_f$ and $\u_w$,
\begin{align}
     \iiiint  q(\f,W,\u_f,\u_w) d\f dW d\u_f d\u_w &= \int_{\f} \int_{\u_f} p(\f|\u_f) q(\u_f) d\f  d\u_f \int_{W} \int_{\u_w} p(W|\u_w) q(\u_w) dW d\u_w \\
     &= \int_{\f}  q(\f) d\f \int_W q(W) dW
\end{align}
where
\begin{align}
q(\f) &= \int \hspace{-0.5mm} p(\f | \u_f) q(\u_f) d\u_f\\
&= \prod_{q=1}^Q \int \N(\f_q | K_{f_q nm} K_{f_q mm}^{-1} \u_{f_q} , \widetilde{K}_{f_q}) \N(\u_{f_q}| \m_{f_q}, \mathbf{S}_{f_q}) d\u_{f_q}\\
 & = \prod_{q=1}^Q \N(\f_q | \bmu_{f_q},\Sigma_{f_q}) \\
q(W)  &= \int \hspace{-0.5mm} p(W | \u_w) q(\u_w) d \u_w\\
&=  \prod_{q,p=1}^{Q,P} \int \N(W_{qp} | K_{w_{qp} nm} K_{w_{qp} mm}^{-1} \u_{w_{qp}} , \widetilde{K}_{w_{qp}}) \N(\u_{w_{qp}}| \m_{w_{qp}}, \mathbf{S}_{w_{qp}}) d\u_{w_{qp}}\\
&= \prod_{q,p=1}^{Q,P} \N(W_{qp} | \bmu_{w_{qp}},\Sigma_{w_{qp}})
\end{align}
with
\begin{align}
\bmu_{f_q} &= K_{f_qnm} K_{f_qmm}^{-1} \m_{f_q}  \\
\bmu_{w_{qp}} &= K_{w_{qp}nm} K_{w_{qp} mm}^{-1} \m_{w_{qp}}  \\
\Sigma_{f_q} &= K_{f_qnn} + K_{f_qnm} K_{f_qmm}^{-1} (\mathbf{S}_{f_q} - K_{f_qmm}) K_{f_qmm}^{-1} K_{f_qmn} \\
\Sigma_{w_{qp}} &= K_{w_{qp}nn} + K_{w_{qp} nm} K_{w_{qp} mm}^{-1} (\mathbf{S}_{w_{qp}} - K_{w_{qp}mm}) K_{w_{qp} mm}^{-1} K_{w_{qp} mn} .
\end{align}

Since the noise term $\boldsymbol\varepsilon$ is assumed to be isotropic Gaussian, density $p(\y|W,\f)$ factorises across all target observations and dimensions. The expectation term in the ELBO then reduces to,
\begin{align}
   \log p(\y ) &\ge \E_{q(W)} \E_{q(\f)} \log p(\y | \f, W) \\
  &= \sum_{i,p=1}^{N,P} \iint \log \N(y_{p,i} |\w_{p,i}^T \f_i, \varepsilon_p^2) q(\f_i,\w_{p,i}) d\w_{p,i} d\f_i - \KL[ q(\u_f, \u_w) || p(\u_f,\u_w)].
\end{align}
The integral with respect to $\f$ can be now solved as
\begin{align}
  \int \log \N(y_{p,i} |\w_{p,i}^T \f_i, \varepsilon_p^2) q(\f_i) d\f_i  &= \log \N(y_{p,i} |\w_{p,i}^T \bmu_{f_i}, \varepsilon_p^2) - \frac{1}{2\varepsilon_p^2} Tr \big[\w_{p,i}^T \Sigma_{f_i} \w_{p,i}\big]\\
  &= \log \N(y_{p,i} |\w_{p,i}^T \bmu_{f_i}, \varepsilon_p^2) - \frac{1}{2\varepsilon_p^2} Tr \big[\Sigma_{f_i} \w_{p,i} \w_{p,i}^T \big]. \numberthis
\end{align}
Next we can marginalize $W$ from the above terms,
\begin{align}
  \int \log \N(y_{p,i} |\w_{p,i}^T \bmu_{f_i}, \varepsilon_p^2)  q(\w_{p,i}) d\w_{p,i} &= \log \N(y_{p,i} |\bmu_{w_{p,i}}^T \bmu_{f_i}, \varepsilon_p^2) - \frac{1}{2\varepsilon_p^2} Tr \big[\bmu_{f_i}^T \Sigma_{w_{q,i}} \bmu_{f_i}\big]\\
  &= \log \N(y_{p,i} |\bmu_{w_{p,i}}^T \bmu_{f_i}, \varepsilon_p^2) - \frac{1}{2\varepsilon_p^2} \sum_{q=1}^{Q} \mu_{f_q,i}^2 \sigma_{w_{qp},i}^2 \\
  \int  \frac{1}{2\varepsilon_p^2} Tr \big[\Sigma_{f_i} \w_{p,i} \w_{p,i}^T \big] q(\w_{p,i}) d\w_{p,i} &=  \frac{1}{2\varepsilon_p^2}Tr \big[ \Sigma_{f_i} (\bmu_{w_{p,i}}\bmu_{w_{p,i}}^T + \Sigma_{w_{q,i}}) \big]\\
  &= \frac{1}{2\varepsilon_p^2} \sum_{q=1}^{Q} \left( \mu_{w_{qp},i}^2 \sigma^2_{f_q,i} + \sigma_{w_{qp},i}^2 \sigma_{f_q,i}^2 \right).
\end{align}

Finally, adding the above results across all $N$ observations and response dimensions $P$ along with Gaussian KL divergence terms, we get the final lowerbound:
\begin{align}
    \log p(\y ) &\ge \sum_{i=1}^{N} \Bigg\{ \sum_{p=1}^{P}  \log \N \Big(y_{p,i} | \sum_{q=1}^Q  \mu_{w_{qp},i} \mu_{f_q,i}, \varepsilon_p^2 \Big) - \frac{1}{2\varepsilon_p^2} \sum_{q,p=1}^{Q,P} \left ( \mu_{w_{qp},i}^2 \sigma^2_{f_q,i} + \mu_{f_q,i}^2 \sigma_{w_{qp},i}^2 + \sigma_{w_{qp},i}^2 \sigma_{f_q,i}^2 \right) \hspace{-1.5mm} \Bigg\}  \\
     &\qquad - \sum_{q,p}^{Q,P} \KL[ q(\u_{w_{qp}},\u_{f_q}) || p(\u_{w_{qp}},\u_{f_q})] \\
     &= \fl_{\textsc{gprn}},
\end{align}
where $\mu_{f_{q},i}$ is the $i$'th element of $\bmu_{f_q}$ and $\sigma_{f_{q},i}^2$ is the $i$'th diagonal element of $\Sigma_{f_q}$ (similarly for $W_{qp}$'s).

\subsection*{C) The stochastic variational bound of the sparse Gaussian process network}

Sparse GPRN is a modification to standard GPRN where sparsity is added to the mixing matrix components. This corresponds to the $p$'th output being a sparse mixture of the latent $Q$ functions, i.e. it can effectively use any subset of the $Q$ latent dimensions by having zeros in the mixing functions. The joint distribution for the model can be written as,
\begin{align}
p(\y, \f, W, \bg) &= p(\y|\f,W) p(\f) p(W | \bg) p(\bg),
\end{align}
where all individual components of latent function $\f$ and mixing matrix $W$ are given GP priors. We encode the sparsity terms $\bg$ for all $Q \times P$ mixing functions $W_{qp}(\x)$ functions as
\begin{align}
p(W_{qp}| \bg_{qp}) &= \N(\w_{qp}| \0, \Phi(\bg_{qp}) \Phi(\bg_{qp})^T \circ K_w ).
\end{align}
To introduce variational inference, the joint model is augmented with three sets of inducing variables for $\f$, $W$ and $\bg$. After marginalizing out the inducing variables similar to SVI for standard GPRN, the lower bund for marginal likelihood can be written as
\begin{align}
    \log p(\y) &\ge \E_{q(\f,W,
    \bg)} \log p(\y | \f, W) - \KL[ q(\u_f,\u_w,\u_g) || p(\u_f,\u_w,\u_g)].
\end{align}
Where the joint distribution in the variational expectation factorizes as $q(\f,W,\bg) = q(\f) q(W|\bg) q(\bg)$. The variational posterior after marginalizing inducing variables is written as,
\begin{align}
q(\f) &= \int \hspace{-0.5mm} q(\f | \u_f) q(\u_f) d\u_f\\
&= \prod_{q=1}^Q \N(\f_q | \bmu_{f_q},\Sigma_{f_q}) \\
q(W)  &= \int \hspace{-0.5mm} q(W | \u_w) q(\u_w) d \u_w\\
&= \prod_{q,p=1}^{Q,P} \N(W_{qp} | \bmu_{w_{qp}},\Sigma_{w_{qp}})\\
q(\bg)  &= \int \hspace{-0.5mm} q(\bg | \u_g) q(\u_g) d \u_g\\
&= \prod_{q,p=1}^{Q,P} \N(g_{qp} | \bmu_{g_{qp}},\Sigma_{g_{qp}})
\end{align}
with
\begin{align}
\bmu_{f_q} &= K_{f_qnm} K_{f_qmm}^{-1} \m_{f_q}  \\
\bmu_{w_{qp}} &= K_{w_{qp}nm} K_{w_{qp} mm}^{-1} \m_{w_{qp}}  \\
\bmu_{g_{qp}} &= K_{g_{qp}nm} K_{g_{qp} mm}^{-1} \m_{g_{qp}}  \\
\Sigma_{f_q} &= K_{f_qnn} + K_{f_qnm} K_{f_qmm}^{-1} (\mathbf{S}_{f_q} - K_{f_qmm}) K_{f_qmm}^{-1} K_{f_qmn} \\
\Sigma_{w_{qp}} &= K_{w_{qp}nn} + K_{w_{qp} nm} K_{w_{qp} mm}^{-1} (\mathbf{S}_{w_{qp}} - K_{w_{qp}mm}) K_{w_{qp} mm}^{-1} K_{w_{qp} mn}\\
\Sigma_{g_{qp}} &= K_{g_{qp}nn} + K_{g_{qp} nm} K_{g_{qp} mm}^{-1} (\mathbf{S}_{g_{qp}} - K_{g_{qp}mm}) K_{g_{qp} mm}^{-1} K_{g_{qp} mn} .
\end{align}

Similar to standard GPRN, with the isotropic Gaussian, density $p(\y|W,\f)$ factorizes across all target observations and dimensions. The expectation term in the ELBO then reduces to
\begin{align}
   \log p(\y ) &\ge \E_{q(W|\bg)} \E_{q(\f)} \E_{q(\bg)} \log p(\y | \f, W) \\
  &= \sum_{i,p=1}^{N,P} \iiint \log \N(y_{p,i} |(\w_{p,i} \circ \bg_{p,i}) ^T \f_i, \varepsilon_p^2) q(\f_i) q(\w_{p,i}|q(\bg_{p,i})q(\bg_{p,i}) d\w_{p,i} d\f_i d\bg_i  - \KL[ q(\u_f, \u_w) || p(\u_f,\u_w)].
\end{align}
The integral with respect to $\f$ can be now solved as
\begin{align}
 \hspace{-10mm} \int \log \N(y_{p,i} |(\w_{p,i} \circ \bg_{p,i})^T \f_i, \varepsilon_p^2) q(\f_i) d\f_i  &= \log \N(y_{p,i} |(\w_{p,i} \circ \bg_{p,i})^T \bmu_{f_i}, \varepsilon_p^2) - \frac{1}{2\varepsilon_p^2} Tr \big[(\w_{p,i} \circ \bg_{p,i})^T \Sigma_{f_i} (\w_{p,i} \circ \bg_{p,i})\big]  \\
  &= \log \N(y_{p,i} |(\w_{p,i} \circ \bg_{p,i})^T \bmu_{f_i}, \varepsilon_p^2) - \frac{1}{2\varepsilon_p^2} Tr \big[\Sigma_{f_i} (\bg_{p,i} \bg_{p,i} ^T \circ \w_{p,i} \w_{p,i}^T) \big].
\end{align}
Next, by integrating individual terms with respect to $W$ we get
\begin{align}
 \hspace{-10mm} \int  \log \N(y_{p,i} |(\w_{p,i} \circ \bg_{p,i})^T \bmu_{f_i}, \varepsilon_p^2)  q(\w_{p,i}) d\w_{p,i} &= \log \N(y_{p,i} |(\bmu_{w_{p,i}} \circ \bg_{p,i})^T \bmu_{f_i}, \varepsilon_p^2) - \frac{1}{2\varepsilon_p^2} Tr \big[\bmu_{f_i}^T (\bg_{p,i}\bg_{p,i}^T \circ \Sigma_{w_{q,i}} ) \bmu_{f_i}\big]  \\
 \hspace{-10mm} \int \frac{1}{2\varepsilon_p^2} Tr \big[\Sigma_{f_i} (\bg_{p,i} \bg_{p,i} ^T \circ \w_{p,i} \w_{p,i}^T) \big] &=  \frac{1}{2\varepsilon_p^2} Tr \big[ \Sigma_{f_i} ( \bg_{p,i}\bg_{p,i}^T \circ (\bmu_{w_{p,i}}\bmu_{w_{p,i}}^T + \Sigma_{w_{q,i}})) \big]
\end{align}
Finally, integrating all the above terms with respect to $\bg$, we get
\begin{align}
  \hspace{-10mm}\int \log \N(y_{p,i} |(\bmu_{w_{p,i}} \circ \bg_{p,i})^T \bmu_{f_i}, \varepsilon_p^2)  q(\bg_{p,i}) d\bg_{p,i} &= \log \N(y_{p,i} |(\bmu_{w_{p,i}} \circ \langle \Phi(\bg_{p,i}) \rangle)^T \bmu_{f_i}, \varepsilon_p^2) \\
  & \qquad - \frac{1}{2\varepsilon_p^2} Tr \big[\bmu_{f_i}^T (\bmu_{w_{p,i}}\bmu_{w_{p,i}}^T \circ Var[\Phi(\bg_{p,i})]) \bmu_{f_i}]\\
  &=  \log \N \Big(y_{p,i} | \sum_{q=1}^Q  \mu_{w_{qp},i} \mu_{g_{qp},i} \mu_{f_q,i}, \varepsilon_p^2 \Big) \\
  & \qquad - \frac{1}{2\varepsilon_p^2} \sum_{q=1}^{Q} \left(\sigma_{g_{qp},i}^2  \mu_{f_q,i}^2 \mu_{w_{qp},i}^2\right)  \numberthis\\
  \hspace{-10mm}\int  \frac{1}{2\varepsilon_p^2} Tr \big[\bmu_{f_i}^T (\bg_{p,i}\bg_{p,i}^T \circ \Sigma_{w_{q,i}} ) \bmu_{f_i}\big]  q(\bg_{p,i}) d\bg_{p,i} &= \frac{1}{2\varepsilon_p^2} Tr \big[\bmu_{f_i}^T\left (( \langle \Phi(\bg_{p,i}) \rangle \langle \Phi(\bg_{p,i}) \rangle^T + Var[\Phi(\bg_{p,i})] )\circ \Sigma_{w_{q,i}} \right) \bmu_{f_i}\big]\\
  & = \frac{1}{2\varepsilon_p^2} \sum_{q=1}^{Q} \left((\mu_{g_{qp},i}^2 + \sigma_{g_{qp},i}^2) \mu_{f_q,i}^2 \sigma_{w_{qp},i}^2 \right)  \numberthis\\
   \hspace{-10mm} \int \frac{1}{2\varepsilon_p^2} Tr \big[ \Sigma_{f_i} ( \bg_{p,i}\bg_{p,i}^T \circ (\bmu_{w_{p,i}}\bmu_{w_{p,i}}^T + \Sigma_{w_{q,i}})) \big]   q(\bg_{p,i}) d\bg_{p,i} &= \frac{1}{2\varepsilon_p^2} Tr \big[ \Sigma_{f_i} \left( ( \langle \Phi(\bg_{p,i}) \rangle \langle \Phi(\bg_{p,i}) \rangle^T + Var[\Phi(\bg_{p,i})] ) \circ \bmu_{w_{p,i}}\bmu_{w_{p,i}}^T \right) \big]  \\
   &\qquad +\frac{1}{2\varepsilon_p^2} Tr \big[ \Sigma_{f_i} \left(( \langle \Phi(\bg_{p,i}) \rangle \langle \Phi(\bg_{p,i}) \rangle^T + Var[\Phi(\bg_{p,i})] ) \circ \Sigma_{w_{q,i}} \right) \big] \\
  & = \frac{1}{2\varepsilon_p^2} \sum_{q=1}^{Q}\left((\mu_{g_{qp},i}^2 + \sigma_{g_{qp},i}^2)  (\mu_{w_{qp},i}^2  \sigma_{f_q,i}^2 + \sigma_{w_{qp},i}^2 \sigma_{f_q,i}^2 ) \right).
\end{align}

Adding above results across all the observations $N$ and output dimensions $P$, we retrieve the final evidence lower bound
\begin{align}
    p(\y) &\ge \sum_{i=1}^{N} \Bigg\{ \sum_{p=1}^{P}  \log \N \Big(y_{p,i} | \sum_{q=1}^Q  \mu_{w_{qp},i} \mu_{g_{qp},i} \mu_{f_q,i}, \varepsilon_p^2 \Big) \\
    & \qquad \qquad - \sum_{q,p=1}^{Q,P} \Big((\mu_{g_{qp},i}^2 + \sigma_{g_{qp},i}^2)  \cdot (\mu_{w_{qp},i}^2  \sigma_{f_q,i}^2  + \mu_{f_q,i}^2 \sigma_{w_{qp},i}^2 + \sigma_{w_{qp},i}^2 \sigma_{f_q,i}^2 ) \Big)  - \sum_{q,p=1}^{Q,P} \left (\sigma_{g_{qp},i}^2  \mu_{f_q,i}^2 \sigma_{w_{qp},i}^2\right) \Bigg\} \\
     &\quad - \sum_{q,p}^{Q,P} \KL[ q(\u_{f_{q}},\u_{w_{qp}},\u_{g_{qp}}) || p(\u_{f_q},\u_{w_{qp}},\u_{g_{qp}})] \\
     &= \fl_{s\textsc{gprn}}.
\end{align}
\end{document}